\documentclass[letterpaper]{article}
\usepackage{aaai}
\usepackage{times}
\usepackage{helvet}
\usepackage{courier}
\usepackage{graphicx}
\usepackage{float}
\begin{document}
%
\title{Emotion Detection in Text: Focusing on Latent Representation}
\author{Armin Seyeditabari \\
	UNC Charlotte  \\
\small	\texttt{sseyedi1@uncc.edu} \normalsize \\\And
	Narges Tabari \\
	University of Virginia \\
\small	\texttt{ns5kn@virginia.edu} \normalsize \\\And
	Shefie Gholizadeh \\
	UNC Charlotte  \\
\small	\texttt{sgholiza@uncc.eduu} \normalsize \\\And
	Wlodek Zadrozny \\
	UNC Charlotte \\
\small	\texttt{wzadrozn@uncc.edu} \normalsize }

\maketitle
\begin{abstract}
\begin{quote}
In recent years, emotion detection in text has become more popular due to its vast potential applications in marketing, political science, psychology, human-computer interaction, artificial intelligence, etc. In this work, we argue that current methods which are based on conventional machine learning models cannot grasp the intricacy of emotional language by ignoring the sequential nature of the text, and the context. These methods, therefore, are not sufficient to create an applicable and generalizable emotion detection methodology. Understanding these limitations, we present a new network based on a bidirectional GRU model to show that capturing more meaningful information from text can significantly improve the performance of these models. The results show significant improvement with an average of 26.8 point increase in F-measure on our test data and 38.6 increase on the totally new dataset. \end{quote}
\end{abstract}

\section{Introduction}

There have been many advances in machine learning methods which help machines understand human behavior better than ever. One of the most important aspects of human behavior is emotion. If machines could detect human emotional expressions, it could be used to improve on verity of applications such as marketing \cite{bagozzi1999role}, human-computer interactions \cite{brave2003emotion}, political science \cite{Druckman2008} etc.

Emotion in humans is complex and hard to distinguish. There have been many emotional models in psychology which tried to classify and point out basic human emotions such as Ekman's 6 basic emotions \cite{ekman1992argument}, Plutchik's wheel of emotions \cite{plutchik1991emotions}, or Parrott's three-level categorization of emotions \cite{parrott2001emotions}. These varieties show that emotions are hard to define, distinguish, and categorize even for human experts.

By adding the complexity of language and the fact that emotion expressions are very complex and context dependant \cite{ben2000subtlety,bazzanella2004emotions,oatley2006understanding}, we can see why detecting emotions in textual data is a challenging task. This difficulty can be seen when human annotators try to assign emotional labels to the text, but using various techniques the annotation task can be accomplished with desirable agreement among the annotators \cite{tafreshi2018sentence}. 

\section{Related Work}
A lot of work has been done on detecting emotion in speech or visual data \cite{han2014speech,lee2015high,wang2018intelligent,zhang2016facial}. But detecting emotions in textual data is a relatively new area that demands more research. There have been many attempts to detect emotions in text using conventional machine learning techniques and handcrafted features in which given the dataset, the authors try to find the best feature set that represents the most and the best information about the text, then passing the converted text as feature vectors to the classifier for training \cite{suttles2013distant,purver2012experimenting,mohammad2012emotional,daume2009frustratingly,roberts2012empatweet,hasan2014emotex,Hasan2018,wang2012harnessing,balabantaray2012multi,wen2014emotion,li2014text,li2015sentence,seyeditabari2018cross}. During the process of creating the feature set, in these methods, some of the most important information in the text such as the sequential nature of the data, and the context will be lost. 

Considering the complexity of the task, and the fact that these models lose a lot of information by using simpler models such as the bag of words model (BOW) or lexicon features, these attempts lead to methods which are not reusable and generalizable. Further improvement in classification algorithms, and trying out new paths is necessary in order to improve the performance of emotion detection methods. Some suggestions that were less present in the literature, are to develop methods that go above lexical representations and consider the flow of the language.  

Due to this sequential nature, recurrent and convolutional neural networks have been used in many NLP tasks and were able to improve the performance in a variety of classification tasks \cite{lai2015recurrent,zhang2015character,zhou2015c,lee2016sequential}. There have been very few works in using deep neural network for emotion detection in text \cite{abdul2017emonet,mundra2017fine}. These models can capture the complexity an context of the language better not only by keeping the sequential information but also by creating hidden representation for the text as a whole and by learning the important features without any additional (and often incomplete) human-designed features.

In this work, we argue that creating a model that can better capture the context and sequential nature of text , can significantly improve the performance in the hard task of emotion detection. We show this by using a recurrent neural network-based classifier that can learn to create a more informative latent representation of the target text as a whole, and we show that this can improve the final performance significantly. Based on that, we suggest focusing on methodologies that increase the quality of these latent representations both contextually and emotionally, can improve the performance of these models. Based on this assumption we propose a deep recurrent neural network architecture to detect discrete emotions in a tweet dataset. The code can be accessed at GitHub [https://github.com/armintabari/Emotion-Detection-RNN].

\section{Experiment}

\subsection{Baseline Approaches}

We compare our approach to two other, the first one uses almost the same tweet data as we use for training, and the second one is the CrowdFlower dataset annotated for emotions. 

In the first one Wang et al. \cite{wang2012harnessing} downloaded over 5M tweets which included one of 131 emotional hashtags based on Parrott's three-level categorization of emotions in seven categories: \textit{joy, sadness, anger, love, fear, thankfulness, surprise}. To assess the quality of using hashtags as labels, the sampled 400 tweets randomly and after comparing human annotations by hashtag labels they came up with simple heuristics to increase the quality of labeling by ignoring tweets with quotations and URLs and only keeping tweets with 5 terms or more that have the emotional hashtags at the end of the tweets. Using these rules they extracted around 2.5M tweets. After sampling another 400 random tweets and comparing it to human annotation the saw that hashtags can classify the tweets with 95\% precision. 
They did some pre-processing by making all words lower-case, replaced user mentions with @user, replaced letters/punctuation that is repeated more than twice with the same two letters/punctuation (e.g., ooooh $\rightarrow{}$ ooh, !!!!! $\rightarrow{}$ !!); normalized some frequently used informal expressions (e.g., ll → will, dnt $\rightarrow{}$ do not); and stripped hash symbols.
They used a sub-sample of their dataset to figure out the best approaches for classification, and after trying two different classifiers (multinomial Naive Bayes and LIBLINEAR) and 12 different feature sets, they got their best results using logistic regression branch for LIBLINEAR classifier and a feature set consist of n-gram(n=1,2), LIWC and MPQA lexicons, WordNet-Affect and POS tags.

In the second one, the reported results are from a paper by \cite{bostan2018analysis} in which they used maximum entropy classifier with bag of words model to classify various emotional datasets. Here we only report part of their result for CrowdFlower dataset that can be mapped to one of our seven labels.

\subsection{Data and preparation}

There are not many free datasets available for emotion classification. Most datasets are subject-specific (i.e. news headlines, fairy tails, etc.) and not big enough to train deep neural networks. Here we use the tweet dataset created by Wang et al. As mentioned in the previous section, they have collected over 2 million tweets by using hashtags for labeling their data. They created a list of words associated with 7 emotions (six emotions from \cite{shaver1987emotion} \textit{love, joy, surprise, anger, sadness fear} plus \textit{thankfulness} (See Table \ref{tab:wangdata}), and used the list as their guide to label the sampled tweets with acceptable quality. 

\begin{table}[h!]
    \centering

\begin{tabular}{l|cr}
     \textbf{Emotion} & \textbf{Hashtags } & \textbf{Number of Tweets}\\ \hline
    joy & 36 & 706,182 \\ 
    sadness & 36 & 616,471 \\ 
    anger & 23 & 574,170 \\ 
    love & 7 & 301,759 \\ 
    fear & 22 & 135,154 \\ 
    thankfulness & 2 & 131,340 \\ 
    surprise & 5 & 23,906 \\ \hline
    \textbf{Total} & \textbf{131} & \textbf{2,488,982} 
\end{tabular}
\caption{Statistics in the original dataset from Wang et al.}
    \label{tab:wangdata}
\end{table}

After pre-processing, they have used 250k tweets as the test set, around 250k as development test and the rest of the data (around 2M) as training data. their best results using LIBLINEAR classifier and a feature set containing n-gram(n=1,2), LIWC and MPQA lexicons, WordNet-Affect and POS tags can be seen in Table \ref{tab:wangresults}. It can be seen that their best results were for high count emotions like \textit{joy} and \textit{sadness} as high as 72.1 in F-measure and worst result was for a low count emotion \textit{surprise} with F-measure of 13.9.

\begin{table}[h!]
    \centering
\begin{tabular}{lr}
     \textbf{Emotion}  & \textbf{F-measure\%}\\ \hline
    joy (28.5\%) & 72.1 \\ 
    sadness (24.6\%) & 64.7\\ 
    anger (23.0\%) & 71.5 \\ 
    love (12.1\%) & 51.5 \\ 
    fear (5.6\%) & 43.9  \\ 
    thankfulness (5.3\%) & 57.1\\ 
    surprise (1.0\%) & 13.9\\ 
\end{tabular}
\caption{Results of final classification in Wang et al.}
    \label{tab:wangresults}
\end{table}

As Twitter is against polishing this many tweets, Wang et al. provided the tweet ids along with their label. For our experiment, we retrieved the tweets in Wang et al.'s dataset by tweet IDs. As the dataset is from 7 years ago We could only download over 1.3 million tweets from around 2.5M tweet IDs in the dataset. The distribution of the data can be seen in Table \ref{tab:ourdata}. 

In our experiment, we used simpler pre-processing steps which will be explained later on in the "Experiment" section.

\begin{table}[h]
    \centering
\begin{tabular}{lr}
     \textbf{Emotion} & \textbf{Number of Tweets}\\ \hline
    joy &  393,631 \\ 
    sadness & 338,015 \\ 
    anger & 298,480 \\ 
    love & 169,267 \\ 
    fear & 73,575 \\ 
    thankfulness & 79,341 \\ 
    surprise & 13,535 \\ \hline
    \textbf{Total} & \textbf{1,387,787} \\ 
\end{tabular}
\caption{Statistics in the downloaded dataset from Wang et al (2012). This is the main dataset used for training the model.}
    \label{tab:ourdata}
\end{table}

\begin{figure*}[t]
    \centering
    \includegraphics[width=0.9\linewidth]{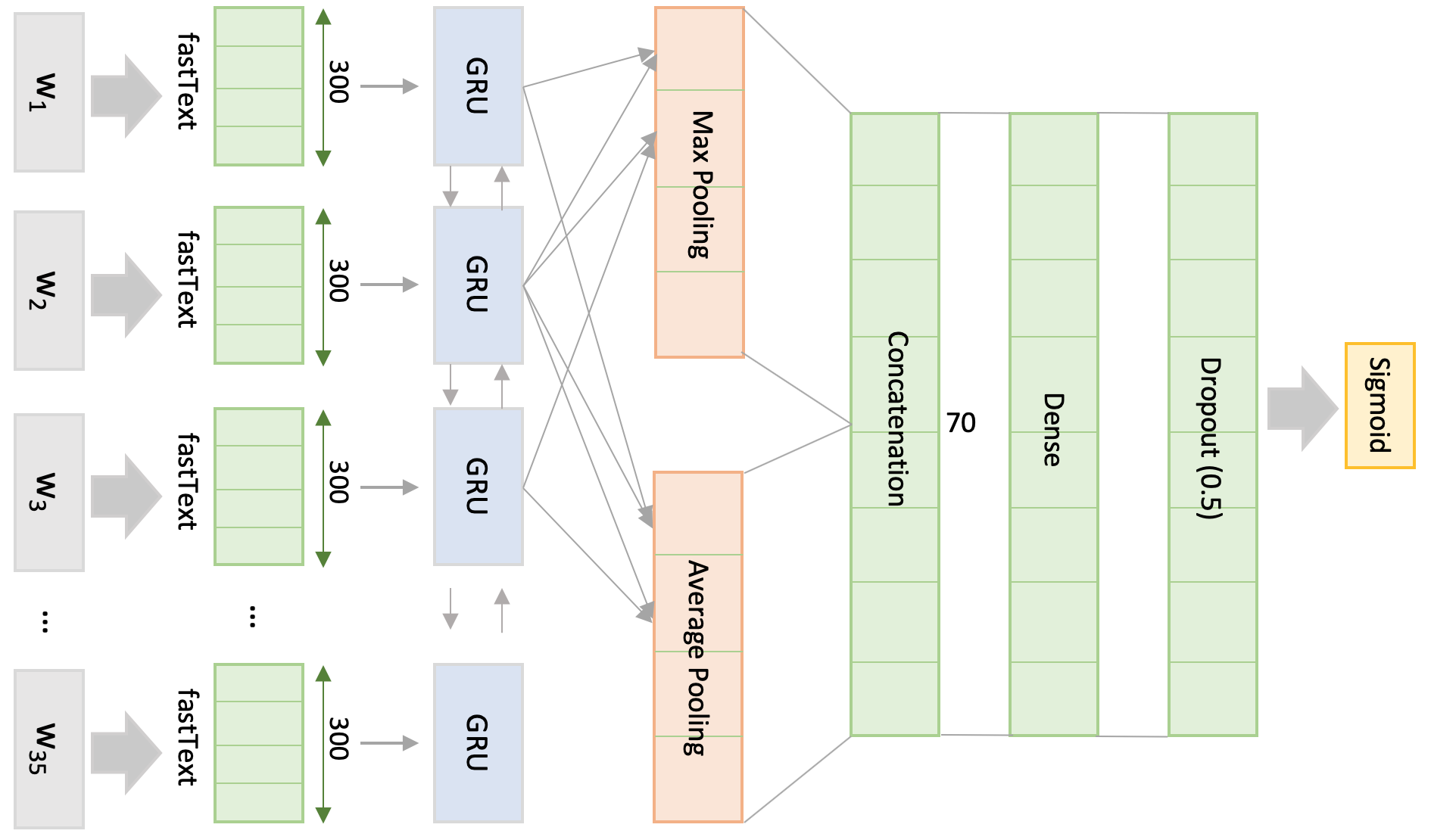}
    \caption{Bidirectional GRU architecture used in our experiment.}
    \label{fig:GRU}
\end{figure*}

\subsection{Model}

In this section, we introduce the deep neural network architecture that we used to classify emotions in the tweets dataset. Emotional expressions are more complex and context-dependent even compared to other forms of expressions based mostly on the complexity and ambiguity of human emotions and emotional expressions and the huge impact of context on the understanding of the expressed emotion. These complexities are what led us to believe lexicon-based features like is normally used in conventional machine learning approaches are unable to capture the intricacy of emotional expressions.  

Our architecture was designed to show that using a model that captures better information about the context and sequential nature of the text can outperform lexicon-based methods commonly used in the literature. As mentioned in the Introduction, Recurrent Neural Networks (RNNs) have been shown to perform well for the verity of tasks in NLP, especially classification tasks. And as our goal was to capture more information about the context and sequential nature of the text, we decided to use a model based on bidirectional RNN, specifically a bidirectional GRU network to analyze the tweets.

For building the emotion classifier, we have decided to use 7 binary classifiers-one for each emotion- each of which uses the same architecture for detecting a specific emotion. You can see the plot diagram of the model in Figure \ref{fig:GRU}. The first layer consists of an embedding lookup layer that will not change during training and will be used to convert each term to its corresponding embedding vector. In our experiments, we tried various word embedding models but saw little difference in their performance. Here we report the results for two which had the best performance among all, ConceptNet Numberbatch \cite{speer2017conceptnet} and fastText \cite{mikolov2018advances} both had 300 dimensions. 

As none of our tweets had more than 35 terms, we set the size of the embedding layer to 35  and added padding to shorter tweets. The output of this layer goes to a bidirectional GRU layer selected to capture the entirety of each tweet before passing its output forward. The goal is to create an intermediate representation for the tweets that capture the sequential nature of the data. For the next step, we use a concatenation of global max-pooling and average-pooling layers (with a window size of two). Then a max-pooling was used to extract the most important features form the GRU output and an average-pooling layer was used to considers all features to create a representation for the text as a whole. These partial representations are then were concatenated to create out final hidden representation. For classification, the output of the concatenation is passed to a dense classification layer with 70 nodes along with a dropout layer with a rate of 50\% to prevent over-fitting. The final layer is a sigmoid layer that generates the final output of the classifier returning the class probability.

\subsection{Experiment}
\label{sec:results}
Minimal pre-processing was done by converting text to lower case, removing the hashtags at the end of tweets and separating each punctuation from the connected token (e.g., awesome!! $\rightarrow{}$ awesome !!) and replacing comma and new-line characters with white space. The text, then, was tokenized using TensorFlow-Keras tokenizer. Top \textit{N} terms were selected and added to our dictionary where N=100k for higher count emotions \textit{joy, sadness, anger, love} and N=50k for \textit{thankfulness and fear} and N=25k for \textit{surprise}. Seven binary classifiers were trained for the seven emotions with a batch size of 250 and for 20 epochs with binary cross-entropy as the objective function and Adam optimizer. The architecture of the model can be seen in Figure \ref{fig:GRU}. For training each classifier, a balanced dataset was created with selecting all tweets from the target set as class 1 and a random sample of the same size from other classes as class 0. For each classifier, 80\% of the data was randomly selected as the training set, and 10\% for the validation set, and 10\% as the test set. As mentioned before we used the two embedding models, ConceptNet Numberbatch  and fastText as the two more modern pre-trained word vector spaces to see how changing the embedding layer can affect the performance. The result of comparison among different embeddings can be seen in Table \ref{tab:embeddingcomparision}. It can be seen that the best performance was divided between the two embedding models with minor performance variations. 

The comparison of our result with Wang et al. can be seen in Table \ref{tab:comparision}. as shown, the results from our model shows significant improvement from 10\% increase in F-measure for a high count emotion \textit{joy} up to 61.7 point increase in F-measure for a low count emotion \textit{surprise}. on average we showed 26.8 point increase in F-measure for all categories and more interestingly our result shows very little variance between different emotions compare to results reported by Wang et al.

\begin{table}[h!]
\centering
\begin{tabular}{l|ccc}
     \textbf{Emotion} & \textbf{Wang et al\%.} & \textbf{Ours\%} & \textbf{Difference\%}\\ \hline
    joy &  72.1 & 82.1 & 10.0\\ 
    sadness & 64.7  & 79.2 & 14.5 \\ 
    anger & 71.5  & 83.7 & 12.2 \\ 
    love & 51.5 & 80.3 & 28.8 \\ 
    fear & 43.9 & 78.1 & 34.2 \\ 
    thankfulness & 57.1 & 83.6 & 26.5 \\ 
    surprise &  13.9 & 75.6 & 61.7 \\ \hline
    Average & 53.5 & 80.4 & 26.8 \\
\end{tabular}
\caption{Results of classification using bidirectional GRU. Reported numbers are F1-measures.}
    \label{tab:comparision}
\end{table}

\begin{table}[h!]
\centering
\begin{tabular}{l|cc}
     \textbf{Emotion} & \textbf{Numberbatch} & \textbf{fastText}\\ \hline
    joy & \textbf{82.11}   & 81.90  \\ 
    sadness & \textbf{79.17}  & 78.71  \\ 
    anger  & 83.44  & \textbf{83.74} \\ 
    love & 79.83  & \textbf{80.29}  \\ 
    fear & 77.61 & \textbf{78.11}  \\ 
    thankfulness  & \textbf{83.64}  & 83.58   \\ 
    surprise  & 75.40  & \textbf{75.58}   \\ 
\end{tabular}
\caption{Results of classification using two embedding models and bidirectional GRU. No meaningful differences was seen between the two models. Reported numbers are F1-measures.}
    \label{tab:embeddingcomparision}
\end{table}


\subsection{Model Performances on New Dataset}

To asses the performance of these models on a totally unseen data, we tried to classify the CrowdFlower emotional tweets dataset. The CrowdFlower dataset consists of 40k tweets annotated via crowd-sourcing each with a single emotional label. This dataset is considered a hard dataset to classify with a lot of noise. The distribution of the dataset can be seen in Table \ref{tab:CF}. The labeling on this dataset is non-standard, so we used the following mapping for labels:
\begin{itemize}
    \item sadness $\rightarrow{}$ sadness
    \item worry $\rightarrow{}$ fear
    \item happiness $\rightarrow{}$ joy
    \item love $\rightarrow{}$ love
    \item surprise $\rightarrow{}$ surprise
    \item anger $\rightarrow{}$ anger
\end{itemize}

We then classified emotions using the pre-trained models and emotionally fitted fastText embedding.  The result can be seen in Table \ref{tab:CFcompare}. The baseline results are from  \cite{bostan2018analysis} done using BOW model and maximum entropy classifier. We saw a huge improvement from 26 point increase in F-measure for the emotion \textit{joy (happiness)} up to 57 point increase for \textit{surprise} with total average increase of 38.6 points. Bostan and Klinger did not report classification results for the emotion \textit{love} so we did not include it in the average. These results show that our trained models perform exceptionally on a totally new dataset with a different method of annotation.

\begin{table}[h!]
    \centering

\begin{tabular}{lr}

     \textbf{Emotion} & \textbf{Number of Tweets}\\ \hline
    neutral &  8638 \\ 
    worry & 8459 \\ 
    happiness & 5209 \\ 
    sadness & 5165 \\ 
    love & 3842 \\ 
    surprise & 2187 \\ 
    fun & 1776 \\ 
    relief & 1526 \\ 
    hate & 1323 \\ 
    enthusiasm & 759 \\ 
    boredom & 179 \\ 
    anger & 110 \\ 
    empty & 827 \\ \hline
    \textbf{Total} & \textbf{40000} \\

\end{tabular}

\caption{Distribution of labels in CrowdFlower dataset.}
    \label{tab:CF}
\end{table}

\begin{table}[h!]
    \centering

\begin{tabular}{l|ccc}

     \textbf{Emotion} & \textbf{Baseline}\% & \textbf{Our Model}\% & \textbf{Difference} \\ \hline
    joy (happiness) &  38 & 64 & 26  \\ 
    sadness & 27  & 65 & 38\\ 
    anger & 24 & 62 & 38 \\ 
    love & - & 66  & - \\ 
    fear (worry) & 31 & 65 & 34 \\ 
    surprise &  9 & 66 & 57\\ \hline
    \textbf{Average} & \textbf{25.8} & \textbf{63.2} & \textbf{38.6} \\

\end{tabular}

\caption{Results from classifying CrowdFlower data using pre-trained model. Reported numbers are F1-measure.}
    \label{tab:CFcompare}
\end{table}

\section{Conclusion and Future Work}
In this paper, we have shown that using the designed RNN based network we could increase the performance of classification dramatically. We showed that keeping the sequential nature of the data can be hugely beneficial when working with textual data especially faced with the hard task of detecting more complex phenomena like emotions. We accomplished that by using a recurrent network in the process of generating our hidden representation. We have also used a max-pooling layer to capture the most relevant features and an average pooling layer to capture the text as a whole proving that we can achieve better performance by focusing on creating a more informative hidden representation. In future we can focus on improving these representations for example by using attention networks \cite{bahdanau2014neural,yang2016hierarchical} to capture a more contextual representation or using language model based methods like BERT \cite{devlin2018bert} that has been shown very successful in various NLP tasks.

\bibliography{refs} 
\bibliographystyle{aaai}

\end{document}